\documentclass[11pt]{article}

\usepackage[preprint]{acl}

\usepackage{times}
\usepackage{latexsym}
\usepackage{tabularx}
\usepackage[table]{xcolor}
\usepackage[T1]{fontenc}
\usepackage[utf8]{inputenc}
\usepackage{microtype}
\usepackage{inconsolata}

\usepackage{url}
\usepackage{booktabs}
\usepackage{amsfonts}
\usepackage{nicefrac}
\usepackage{xurl}
\usepackage{enumitem}
\usepackage{multicol}
\usepackage{CJKutf8}
\usepackage{amsmath}
\usepackage{siunitx}
\usepackage{floatflt}
\usepackage{graphicx}
\usepackage{wrapfig}
\usepackage{multirow}
\usepackage{pifont}
\usepackage{amssymb}
\usepackage{makecell}
\usepackage[colorinlistoftodos]{todonotes}
\usepackage{array}
\usepackage{mathtools}
\usepackage{amsthm}
\usepackage{ragged2e}
\usepackage[capitalize,noabbrev]{cleveref}
\usepackage{adjustbox}
\usepackage{caption}
\usepackage{arydshln}
\usepackage{fontawesome}
\usepackage{tcolorbox}
\tcbuselibrary{skins,breakable}
\usepackage{soulutf8}
\usepackage{soul}
\usepackage{algorithm}
\usepackage{algorithmicx}
\usepackage{algpseudocode}
\usepackage{subfigure}

\definecolor{srconebg}{HTML}{DFF3E4}
\definecolor{srctwobg}{HTML}{DCEBFF}
\definecolor{srcthreebg}{HTML}{EADCF8}

\newcommand{\hlone}[1]{{\sethlcolor{srconebg}\hl{#1}}}
\newcommand{\hltwo}[1]{{\sethlcolor{srctwobg}\hl{#1}}}
\newcommand{\hlthree}[1]{{\sethlcolor{srcthreebg}\hl{#1}}}

\definecolor{tred}{RGB}{251, 130, 132}
\definecolor{torange}{RGB}{247, 162, 116}
\definecolor{tyellow}{RGB}{251, 218, 140}
\definecolor{tgreen}{RGB}{127, 204, 181}
\definecolor{tblue}{RGB}{89, 177, 215}
\definecolor{insightblue}{RGB}{162, 210, 255}
\definecolor{questionred}{RGB}{255, 175, 204}
\definecolor{MintLight}{RGB}{223,243,228}
\definecolor{MintDark}{RGB}{80,160,110}
\definecolor{SkyLight}{RGB}{220,235,255}
\definecolor{SkyDark}{RGB}{60,130,190}
\definecolor{LavenderLight}{RGB}{217,203,255}
\definecolor{LavenderDark}{RGB}{130,90,190}
\definecolor{PeachLight}{RGB}{255,220,180}
\definecolor{PeachDark}{RGB}{190,120,50}
\definecolor{SageLight}{RGB}{200,230,200}
\definecolor{SageDark}{RGB}{70,140,80}
\definecolor{TableHeader}{RGB}{240,240,240}

\title{\textbf{\textit{SciResearcher}}: Scaling Deep Research Agents for\\
Frontier Scientific Reasoning}

\author{Tianshi Zheng$^1$, Rui Wang$^2$, Xiyun Li$^3$, Kelvin Kiu-Wai Tam$^1$, Newt Hue-Nam K. Nguyen$^1$\\ \textbf{Wei Fan$^1$, Yangqiu Song$^1$, Tianqing Fang$^3$}\\
$^1$HKUST, 
$^2$CUHK,
$^3$Tencent AI Lab\\
\texttt{tzhengad@connect.ust.hk, fangtq229@gmail.com}\\}

\begin{document}

\maketitle

\begin{abstract}
Frontier scientific reasoning is rapidly emerging as a key foundation for advancing AI agents in automated scientific discovery. Deep research agents offer a promising approach to this challenge. These models develop robust problem-solving capabilities through post-training on information-seeking tasks, which are typically curated via knowledge graph construction or iterative web browsing. However, these strategies face inherent limitations in frontier science, where domain-specific knowledge is scattered across sparse and heterogeneous academic sources, and problem solving requires sophisticated computation and reasoning far beyond factual recall. To bridge this gap, we introduce \textbf{\textit{SciResearcher}}, a fully automated agentic framework for frontier-science data construction. SciResearcher synthesizes diverse conceptual and computational tasks grounded in academic evidence, while eliciting information acquisition, tool-integrated reasoning, and long-horizon capabilities. Leveraging the curated data for supervised fine-tuning and agentic reinforcement learning, we develop \textbf{\textit{SciResearcher}}-8B, an agent foundation model that achieves 19.46\% on the HLE-Bio/Chem-Gold benchmark, establishing a new state of the art at its parameter scale and surpassing several larger proprietary agents. It further achieves 13--15\% absolute gains on SuperGPQA-Hard-Biology and TRQA-Literature benchmarks. Overall, SciResearcher introduces a new paradigm for automated data construction for frontier scientific reasoning and offers a scalable path toward future scientific agents.
\end{abstract}

\begin{figure}[t]
    \centering
    \vspace{-0.3cm}
    \includegraphics[width=\columnwidth]{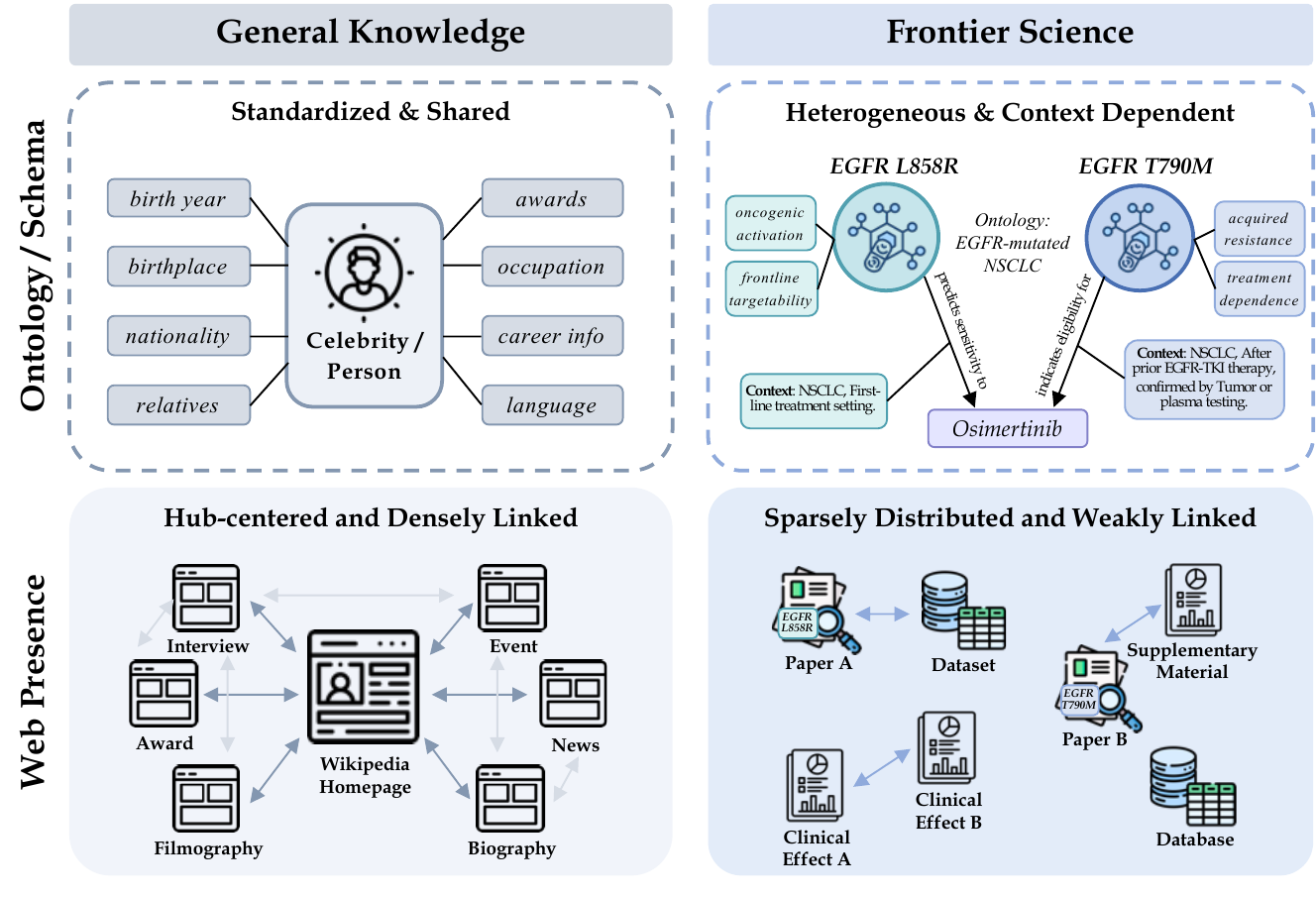}
    \caption{Comparison of ontology and web presence between general knowledge and frontier science.}
    \vspace{-0.3cm}
    \label{fig:egfr_context_schema}
\end{figure}
\section{Introduction}

Frontier scientific reasoning captures an AI system’s ability to solve challenging, expert-level scientific problems at the boundaries of human knowledge \citep{hle,wang2026frontierscienceevaluatingaisability}. Such problems arise in scientific domains where relevant knowledge is often incomplete, rapidly evolving, and distributed across diverse sources \citep{gottweis2025aicoscientist,corywright2024aihilbert}. As AI-driven scientific discovery systems such as LLM-powered research assistants~\citep{schmidgall2025agentlaboratoryusingllm,si2024llmsgeneratenovelresearch} and autonomous AI scientists~\citep{lu2024aiscientistfullyautomated,mitchener2025kosmosaiscientistautonomous,zheng-etal-2025-automation} become increasingly prevalent and capable, reasoning effectively and reliably in these applications is becoming critical.

Deep research agents \citep{openai2025deepresearch,citron2024gemini_deep_research} have emerged as a promising approach to frontier scientific reasoning, benefiting from their ability to acquire up-to-date knowledge through real-time web search and to carry out long-horizon tasks. A central strategy for advancing deep research agents is agent post-training on information-seeking tasks. For general factual knowledge, automated construction of information-seeking tasks largely follows two paradigms: (1) knowledge graph construction-based methods \citep{websailor,tao2025webshaperagenticallydatasynthesizing,li2025websailorv2bridgingchasmproprietary}, which first build a structured graph over entities or webpages and then sample paths or subgraphs to synthesize tasks; and (2) agent-based methods \citep{webdancer,wu2025webwalkerbenchmarkingllmsweb,liu2025webexplorerexploreevolvetraining}, which instead allow an agent to iteratively search and browse from seed entities or URLs to construct a local information space for task synthesis. Existing instantiations of both paradigms, however, are largely grounded in Wikipedia-like, entity-centric factual knowledge, and many of their constructed tasks emphasize entity-, attribute-, or fact-seeking supervision.

While these paradigms have achieved strong results for general-domain information seeking, they are inherently limited in their applicability to frontier scientific reasoning (Figure~\ref{fig:egfr_context_schema}). First, frontier scientific entities are associated with heterogeneous, noisy, and highly context-dependent ontologies \citep{zhang-etal-2021-fine,10.3389/fpls.2023.1279694}. Unlike general-domain entities, which are often described by relatively standardized attributes, scientific entities rarely admit a shared attribute schema, and even similar attributes require substantial domain-specific context to be meaningful. This challenges the feasibility of structured graph construction. Second, knowledge relevant to frontier scientific problems is often sparse and fragmented across loosely connected academic sources, rather than organized as densely interlinked webpages \citep{baulin2025discoveryengineframeworkaidriven,shen-etal-2018-web}. As a result, frontier scientific concepts rarely have a canonical entry point analogous to a Wikipedia page, making continuous web traversal difficult to realize effectively. Moreover, frontier scientific reasoning often requires nontrivial computation over complex scenarios and scientific models \citep{hle,wang2026frontierscienceevaluatingaisability,mudur2025feabenchevaluatinglanguagemodels}, which is not captured by deep research task construction approaches.

To address these limitations, we introduce \textit{\textbf{SciResearcher}}, an automated data construction framework for frontier scientific reasoning. The framework operates over heterogeneous scientific concepts, integrates evidence scattered across weakly connected sources, and treats scientific computation as a core component of reasoning. Built on a curated pool of frontier scientific entities, it comprises two data construction pipelines, targeting \textit{conceptual} and \textit{computational} questions, respectively. For conceptual task construction, we employ web agents to browse from seed scientific entities, gather academic evidence, and generate rich, grounded questions. We then apply an iterative anchor\footnote{We define an \textit{anchor} as the key, decisive scientific entity that plays a central role in a question, serving as its signature referent and the primary handle for reasoning about the problem.}-based task augmentation stage to increase task complexity. For computational task construction, we perform a three-level evidence selection process to identify novel and challenging computational models (typically comprising governing equations, mechanistic understanding, and required input parameters) associated with seed entities, and then generate questions and obtain reference answers through voting-based solver verification. Leveraging this framework, we construct SciResearcherQA, a challenging scientific reasoning dataset that stress-tests long-horizon, multi-evidence problem solving grounded in frontier knowledge and computation.

Using SciResearcherQA, we perform agent post-training following an established recipe that begins with supervised fine-tuning using rejection sampling and followed by reinforcement learning via GRPO \citep{shao2024deepseekmathpushinglimitsmathematical}. Based on Qwen3-8B \citep{yang2025qwen3technicalreport}, we train \textit{\textbf{SciResearcher-8B}}, which achieves strong performance across multiple frontier scientific reasoning benchmarks, including HLE-Bio/Chem-Gold \citep{white2025hlewrong}, SuperGPQA-Hard-Biology \citep{pteam2025supergpqascalingllmevaluation}, and TRQA-Literature \citep{Zhang2025.06.03.657658}. On HLE-Bio/Chem-Gold, our agent attains 19.46\% pass@1 and 31.54\% pass@3, outperforming existing scientific agents and deep research agents that rely on larger LLM backbones, and approaching the performance of leading proprietary deep research systems such as OpenAI Deep Research \citep{openai2025deepresearch}. Moreover, SciResearcher-8B yields substantial gains on SuperGPQA-Bio-Hard and TRQA-Literature, improving absolute performance by 13.04\% and 14.54\%, respectively.

Further analysis shows that post-training leads to substantially more extended and tool-intensive reasoning behavior: compared with the baseline, our agent produces trajectories that are 0.3--2.7$\times$ longer and exhibits correspondingly higher tool-use frequency. Interestingly, reinforcement learning induces an adaptive effect relative to the SFT checkpoint: on the more challenging HLE-Bio/Chem-Gold benchmark, the agent learns to allocate more steps and tool calls, whereas on relatively easier benchmarks it uses slightly fewer steps while still achieving stronger performance. These results suggest that SciResearcherQA improves frontier scientific reasoning performance together with cultivating more adaptive long-horizon information-seeking behavior.

Overall, SciResearcher offers a new perspective on automated data construction, extending it from general information-seeking tasks to frontier scientific reasoning. Our results highlight the promise of automatically constructed scientific training data for scaling deep research agents toward more capable and reliable scientific problem solving. All resources will be released upon acceptance.

\section{The \textit{SciResearcher} Data Construction Framework}

\begin{figure*}[t]
    \centering
    \includegraphics[width=\textwidth]{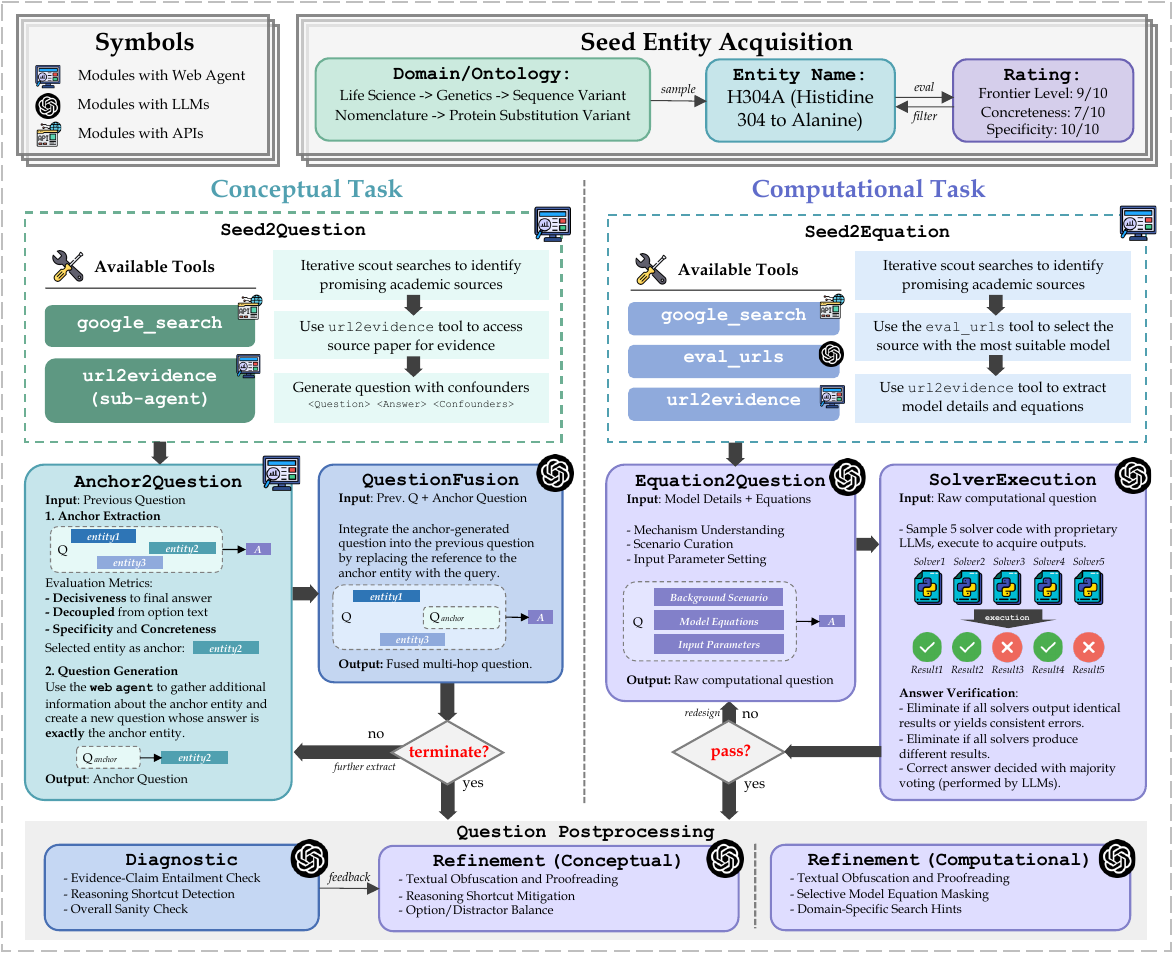}
    \caption{Overview of our \textit{\textbf{SciResearcher}} data construction framework.}
    \label{fig:main_framework}
    \vspace{-0.3cm}
\end{figure*}

In this section, we introduce \textit{\textbf{SciResearcher}}, a fully automated data construction framework (Figure~\ref{fig:main_framework}) for frontier scientific reasoning tasks. Given a curated set of seed entities, SciResearcher generates both conceptual and computational reasoning tasks grounded in academic evidence through agentic web exploration. Examples of the generated questions are shown in Table~\ref{tab:case_studies}.

\subsection{Seed Entity Acquisition}
High-quality, domain-specific scientific entities are essential for constructing frontier scientific tasks. To obtain them, we implement a three-stage seed entity acquisition pipeline. First, we curate a pool of scientific ontologies in biology and chemistry with domain-specific annotations. Second, we use LLMs to generate candidate entities based on these curated ontologies, thereby constructing an entity pool for each scientific domain. Third, we automatically assess all entities using three metrics—frontier relevance, concreteness, and specificity—to evaluate their quality. Finally, we collect the resulting high-quality entities as our seed entity set for task curation.

\subsection{Conceptual Task Curation}
Starting from a target seed entity, we employ a web agent with a proprietary LLM backbone to iteratively search for and browse academic sources on the web, with the goal of constructing a base conceptual question grounded in verifiable scientific evidence. Concretely, the agent first performs iterative scout searches to identify promising academic sources relevant to the seed entity. It then uses a \texttt{url2evidence} sub-agent to access the selected paper, extract the key supporting evidence, and formulate an initial conceptual multiple-choice question together with plausible confounders. This initial question serves as the semantic backbone for subsequent augmentation.

\begin{figure*}[t]
    \centering
    \includegraphics[width=\textwidth]{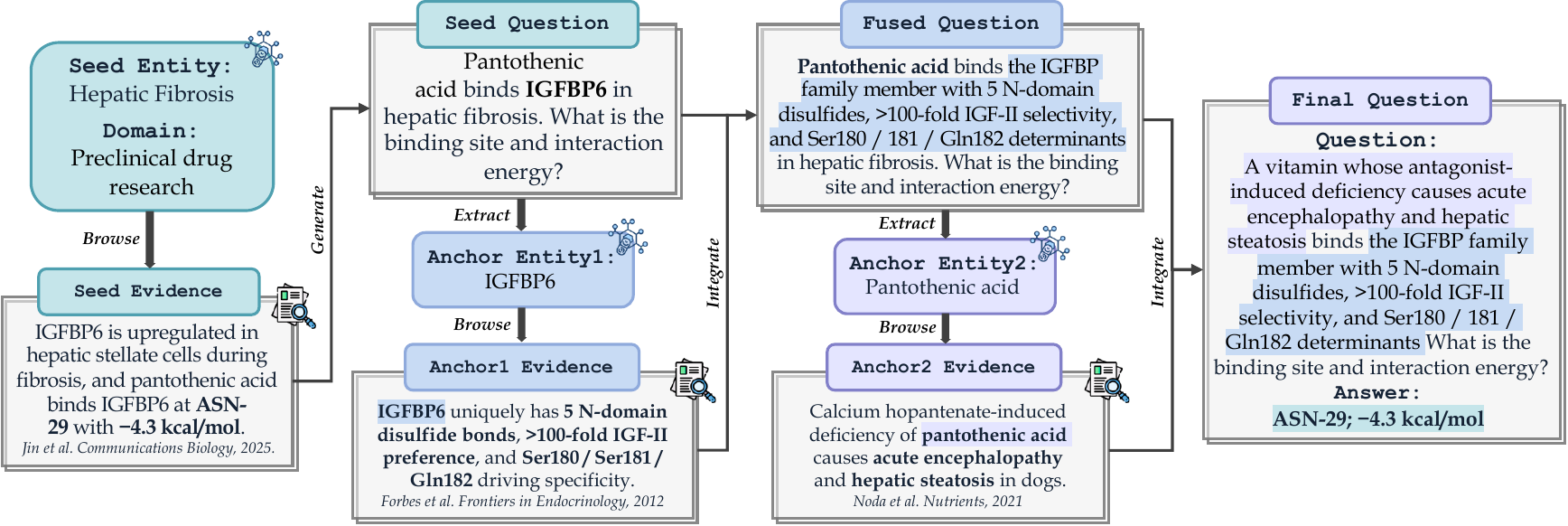}
    \caption{A running example of a question evolution pipeline for conceptual task curation. Question fusion and postprocessing details are omitted.}
    \label{fig:running_example}
    \vspace{-0.3cm}
\end{figure*}

To increase task complexity beyond single-hop retrieval, we further perform anchor-based question augmentation. Given the current question, we first extract candidate anchor entities from the question text and evaluate them using three criteria: whether the entity is decisive for deriving the final answer, whether it is decoupled from the surface form of the answer options, and whether it is sufficiently specific and concrete to support further evidence-grounded expansion. After selecting the best anchor, we invoke a new web agent instance to gather additional academic evidence about that anchor and generate a new question whose answer is exactly the anchor entity. This newly generated question is then fused back into the previous question by replacing the original anchor mention, thereby converting a direct clue into an additional reasoning step. The augmentation process can be repeated multiple times, recursively transforming a seed question into a multi-hop question that requires long-horizon browsing and evidence aggregation across multiple independent sources. Figure~\ref{fig:running_example} shows a running example of this question evolution process.

Unlike prior agent-based approaches such as WebDancer~\citep{webdancer} and WebExplorer~\citep{liu2025webexplorerexploreevolvetraining}, our conceptual task curation instantiates each augmentation step with a separate web agent instance. This design reduces dependence on a single search trajectory and more explicitly stress-tests an agent's ability to perform multi-source academic retrieval, cross-source integration, and compositional reasoning.

\subsection{Computational Task Curation}
In contrast to conceptual tasks, which primarily evaluate information seeking and synthesis, computational tasks additionally require agents to apply retrieved scientific knowledge to perform nontrivial quantitative reasoning.

The pipeline begins with a tailored web agent that identifies and extracts the most appropriate advanced computational model associated with the seed entity through a three-level evidence selection process. First, multiple scouting searches are performed to identify promising sources based on their titles and content snippets. Second, the selected links are evaluated and filtered using the \texttt{eval\_urls} tool, which applies four metrics---model exclusiveness, search identifiability, computational complexity, and LLM unfamiliarity---to support comprehensive assessment. Third, sub-agents are deployed to conduct a deep dive into the final selected URLs, extracting the complete model specification together with the scenarios and constraints required for its application.

Based on the extracted model, the system constructs a scenario-based computational question. This process requires understanding the scientific mechanism encoded by the model, curating a realistic background scenario, and specifying the necessary input parameters. The resulting question therefore tests whether an agent can not only retrieve the relevant scientific source, but also instantiate the model correctly in a concrete setting.

Because such generated questions do not come with a guaranteed correct answer, we further perform answer acquisition and verification. Specifically, we sample five candidate Python solvers from proprietary LLMs and execute them to obtain candidate outputs. We then filter out low-quality questions based on solver agreement patterns: questions are rejected if all solvers return the same result, if all solvers yield consistent errors, or if all solvers produce different answers. These cases typically indicate that the question is respectively too trivial, not executable, or too unstable. For the remaining questions, the final answer is determined by majority voting over solver outputs, followed by LLM-based verification. Questions that fail this verification process are sent back for redesign.

\subsection{Question Postprocessing}
After generating raw questions, we apply a postprocessing stage to improve question quality, reduce shortcut-based answering, and increase the amount of retrieval and reasoning required to solve each task. We begin with a general diagnostic pass, which includes evidence--claim entailment checking, reasoning shortcut detection, and overall sanity checking. Based on diagnostic feedback, we then apply category-specific refinement. For conceptual questions, we perform textual obfuscation and proofreading, mitigate reasoning shortcuts, and improve the balance of answer options and distractors. For computational questions, we additionally perform selective masking of model equations and inject domain-specific search hints to encourage agents to retrieve the relevant scientific source and reconstruct the appropriate computational model.

\subsection{Analysis on SciResearcherQA}

\begin{figure*}[t]
    \centering
    \includegraphics[width=\textwidth]{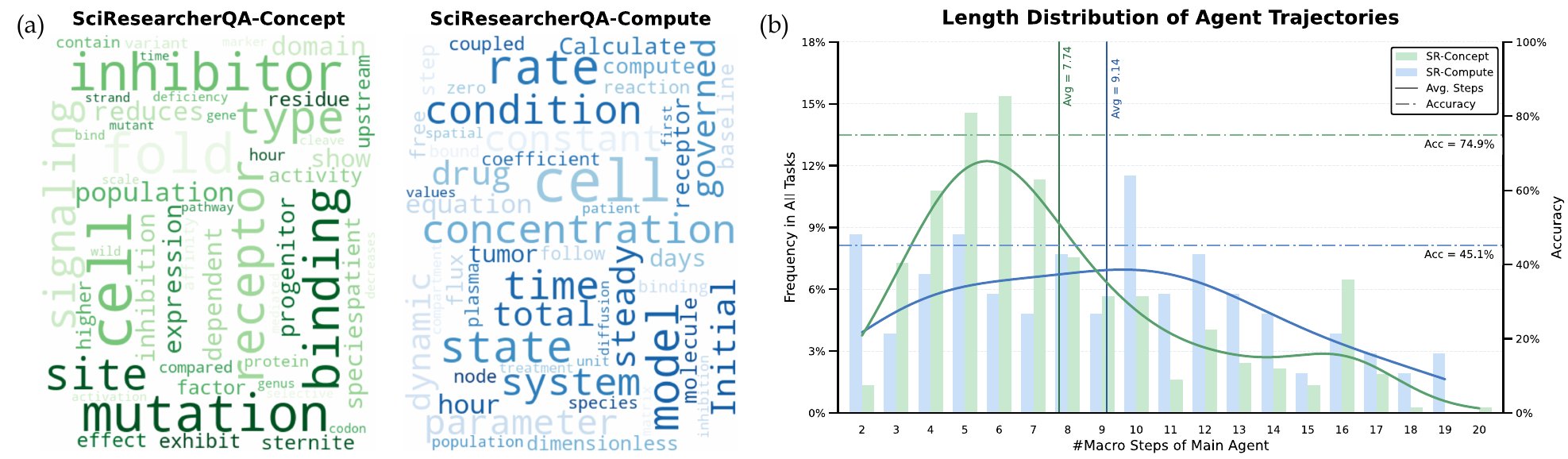}
    \caption{(a) Word clouds of the curated questions from the two pipelines. (b) Distribution and performance of Claude-Sonnet-4.5 across the two question types.}
    \label{fig:data_analysis}
\vspace{-0.3cm}

\end{figure*}

Figure~\ref{fig:data_analysis}(a) visualizes the most frequent words in the two question types of SciResearcherQA. Both word clouds contain domain-specific terminology from biology and chemistry, such as \textit{cell}, \textit{reaction}, and \textit{drug}. Conceptual questions tend to emphasize more qualitative terms, such as \textit{effect}, \textit{dependent}, and \textit{deficiency}, whereas computational questions feature more quantitative terms, such as \textit{rate}, \textit{concentration}, and \textit{parameter}.

Figure~\ref{fig:data_analysis}(b) further presents the trajectory distribution of an agent using Claude-Sonnet-4.5 as the backbone. On conceptual questions, the agent achieved an accuracy of 74.9\% with an average of 7.74 macro steps\footnote{Since we use the multi-agent framework of Cognitive Kernel-Pro, a macro step denotes one planning step and one action step of the main agent. On average, one macro step corresponds to 4.1--4.9 total LLM calls in our experiments.}, whereas on computational questions it achieved 45.1\% accuracy with an average of 9.14 macro steps. These results highlight both the substantial difficulty of the benchmark and its long-horizon reasoning requirements. Compared with conceptual questions, computational questions are markedly more challenging for current agents, requiring more reasoning steps on average while yielding substantially lower accuracy.

To further assess data quality and detect potential data leakage, we conducted a human evaluation and dataset overlap analysis on SciResearcherQA, with the results presented in Appendix \ref{app:data_quality_assessment}.

\section{Experiments}
\subsection{Experimental Setups}

\paragraph{Deep Research Agent Framework}
In our experiments, we adopt Cognitive Kernel-Pro~\citep{fang2025cognitivekernelproframeworkdeep} as our agent framework. CK-Pro is a multi-agent framework optimized for GAIA-like~\citep{GAIA} information-seeking tasks, where a main agent coordinates two specialized sub-agents for web browsing and file analysis. We further adapt it at the prompt and instruction levels to better align it with frontier scientific domains, resulting in a 4--8\% performance improvement in preliminary experiments over the original framework.

\paragraph{Benchmarks}
Following prior work~\citep{chai2025scimastergeneralpurposescientificai,tang2025eigen1adaptivemultiagentrefinement}, we evaluate our method on three frontier scientific reasoning benchmarks: 1) \textbf{HLE-Bio/Chem-Gold}~\citep{white2025hlewrong} is an expert-verified subset of Humanity's Last Exam~\citep{hle}, comprising 149 highly challenging questions in advanced biology and chemistry. 2) \textbf{SuperGPQA-Hard-Biology}~\citep{pteam2025supergpqascalingllmevaluation} contains 92 expert-annotated biology questions that emphasize difficult, reasoning-intensive scientific problem solving. 3) \textbf{TRQA-Literature}~\citep{Zhang2025.06.03.657658} is a knowledge-intensive benchmark grounded in advanced therapeutic research literature. Together, these benchmarks provide complementary coverage of frontier scientific reasoning, ranging from research-level scientific understanding to literature-grounded multi-step inference.

\paragraph{Data and Training}
\label{sec:data_training}

\begin{table}[t]
\centering
\small
\resizebox{\columnwidth}{!}{
\begin{tabular}{lrr}
\toprule
\textbf{Dataset} & \textbf{\# Tasks} & \textbf{\# Steps} \\
\midrule
SciResearcherQA-Concept & 371 & 2,872 \\
SciResearcherQA-Compute & 104 & 951 \\
TRQA-Literature \citep{Zhang2025.06.03.657658} & 172 & 932 \\
SciBench \citep{wang2024scibenchevaluatingcollegelevelscientific} & 80 & 350 \\
\midrule
\textbf{Total} & \textbf{727} & \textbf{5,105} \\
\bottomrule
\end{tabular}
}
\caption{Training data composition. \textbf{\# Tasks} indicates number of QA pairs and \textbf{\# Steps} indicates number of step-level messages for training.}

\label{tab:data_composition}
\vspace{-0.3cm}
\end{table}

The composition of the training data is summarized in Table~\ref{tab:data_composition}. In addition to the two data types generated by SciResearcher, we introduce two auxiliary data sources to improve distributional balance. First, we incorporate a small subset of SciBench~\citep{wang2024scibenchevaluatingcollegelevelscientific} as a source of relatively simple scientific reasoning questions, which helps offset the difficulty of SciResearcherQA-Compute and reduce overthinking during training. Second, we include TRQA as a source of multiple-selection MCQs to counterbalance the predominance of single-selection MCQs in SciResearcherQA-Concept. For evaluations on TRQA-Literature, we train a separate checkpoint with TRQA removed from the training mixture.

We train our agent foundation model, \textbf{\textit{SciResearcher}}-8B, based on Qwen3-8B~\citep{yang2025qwen3technicalreport}, following the standard two-stage training paradigm of cold-start supervised fine-tuning (SFT) followed by reinforcement learning. In the first stage, we collect agent trajectories using Claude-Sonnet-4.5~\citep{anthropic2025claudesonnet45} as the teacher model and perform supervised fine-tuning with rejection sampling to initialize the model's tool-use and long-horizon decision-making abilities. In the second stage, we further optimize the model with reinforcement learning using the GRPO algorithm~\citep{shao2024deepseekmathpushinglimitsmathematical} with outcome-only rewards, encouraging the agent to discover more effective task-completion strategies through interaction.

Our training objective is to improve the main agent's ability to carry out long-horizon scientific tasks, particularly in planning, tool use, and multi-step execution. Accordingly, we train only on trajectories produced by the main agent. The web-browsing and file-analysis sub-agents are kept frozen throughout training and are treated as external tools rather than trainable components.

\begin{table*}[t]

\centering
\small
\setlength{\tabcolsep}{3.5pt}
\renewcommand{\arraystretch}{1.08}
\begin{tabularx}{\textwidth}{l >{\raggedright\arraybackslash}X c c c c}\toprule
\textbf{Agent Framework} & \textbf{LLM Backbone} & \textbf{HLE-Gold} & \textbf{SuperGPQA-Hard} & \textbf{TRQA*} & \textbf{Avg.} \\
\midrule

\multicolumn{6}{l}{\cellcolor[RGB]{223,243,228}\emph{Vanilla LLMs}} \\
\midrule
-- & Qwen3-32B & 5.37 & 31.52 & 37.79 & 24.89 \\
-- & Kimi-K2 & 6.71 & 48.91 & 38.37 & 31.33 \\
-- & Deepseek V3.1 & 13.42 & 66.30 & 43.60 & 41.11 \\
-- & Gemini-2.5 Pro & 18.79 & 65.22 & 45.93 & 43.31 \\
\midrule

\multicolumn{6}{l}{\cellcolor[RGB]{220,235,255}\emph{Proprietary Agents}} \\
\midrule
AutoGen \citep{wu2024autogen} & GPT-4.1 & 7.38 & 29.35 & 51.74 & 29.49 \\
SciMaster \citep{chai2025scimastergeneralpurposescientificai} & GPT-4.1 & 9.45 & 19.78 & 47.67 & 25.63 \\
Biomni \citep{huang2025biomni} & GPT-4.1 & 10.74 & 43.48 & 41.09 & 31.77 \\
OpenAI Deep Research \citep{openai2025deepresearch} & o4-mini & 22.82 & 39.13 & - & - \\
\midrule

\multicolumn{6}{l}{\cellcolor[RGB]{217,203,255}\emph{Open-source Agents}} \\
\midrule
Search-R1 \citep{jin2025searchr1trainingllmsreason} & Search-R1-7B & 8.05 & 17.39 & 17.44 & 14.29 \\
WebThinker \citep{li2025webthinkerempoweringlargereasoning} & WebThinker-R1-7B & 8.72 & 20.65 & 20.35 & 16.57 \\
 & WebThinker-R1-14B & 12.08 & 38.04 & 31.40 & 27.17 \\
WebSailor \citep{websailor} & WebSailor-7B & 10.74 & 14.13 & 45.93 & 23.60 \\
 & WebSailor-32B & 15.44 & 28.26 & 54.07 & 32.59 \\
\midrule

\multirow{5}{*}{\parbox{3.2cm}{\textbf{Cognitive\,Kernel-Pro} \citep{fang2025cognitivekernelproframeworkdeep}}}
& Qwen3-8B & 8.05 & 22.83 & 34.88 & 21.92 \\
& Qwen3-32B & 10.74 & \underline{38.04} & 46.51 & 31.76 \\
& \textit{\textbf{SciResearcher}}-8B-SFT & 12.75 & 31.52 & 47.67 & 30.65 \\
& \textit{\textbf{SciResearcher}}-8B-RL & \underline{19.46} & 35.87 & \underline{49.42} & \underline{34.92} \\
& \; -pass@3 & \textbf{31.54} & \textbf{51.09} & \textbf{60.47} & \textbf{47.70} \\
\bottomrule
\end{tabularx}
\caption{Performance comparison on HLE-Bio/Chem-Gold (n=149), SuperGPQA-Hard-Biology (n=92), and TRQA-Literature (n=172). Results of proprietary agent baselines are reported by \citet{tang2025eigen1adaptivemultiagentrefinement}. *Note: In TRQA experiments, we exclude TRQA from the training data of SciResearcher.}

\label{tab:agent_benchmark_results}
\vspace{-0.3cm}
\end{table*}
\subsection{Results and Analyses}

\paragraph{Main Result}The main experimental results are reported in Table~\ref{tab:agent_benchmark_results}. Overall, \textit{\textbf{SciResearcher}}-8B yields substantial improvements over the base Qwen3-8B agent across all three frontier scientific reasoning benchmarks, demonstrating the effectiveness of our overall approach. On HLE-Bio/Chem-Gold, \textit{\textbf{SciResearcher}}-8B-RL achieves 19.46\%, improving over the baseline by more than 11 absolute points and \textbf{outperforming prior proprietary scientific agents and open-source deep research agents}, all of which use substantially larger backbone models or trained in much larger data scale. On SuperGPQA-Hard-Biology and TRQA-Literature, \textit{\textbf{SciResearcher}}-8B-RL reaches 35.87\% and 49.42\%, corresponding to absolute gains of 13.04 and 14.54 points over the baseline, respectively. These results indicate that our training setup consistently improves performance on both literature-grounded conceptual reasoning and more demanding scientific problem solving. We also observe significant gains under \textit{pass@3} evaluation, suggesting additional headroom for performance improvement through inference-time scaling.

\paragraph{Ablation Study}
Table~\ref{tab:ablation_training_settings} shows the effect of each training data component in the SFT stage. Adding SciResearcherQA-Concept improves performance on both HLE-Gold and SuperGPQA-Hard, and SciResearcherQA-Compute brings further gains, confirming the complementary value of conceptual and computational task curation. The auxiliary TRQA and SciBench data provide additional improvements, suggesting that they help broaden the training distribution. Overall, the ablation confirms that SciResearcherQA is the main driver of the performance gains, while auxiliary data further strengthens robustness.

\begin{figure*}[t]
    \centering
    \includegraphics[width=\textwidth]{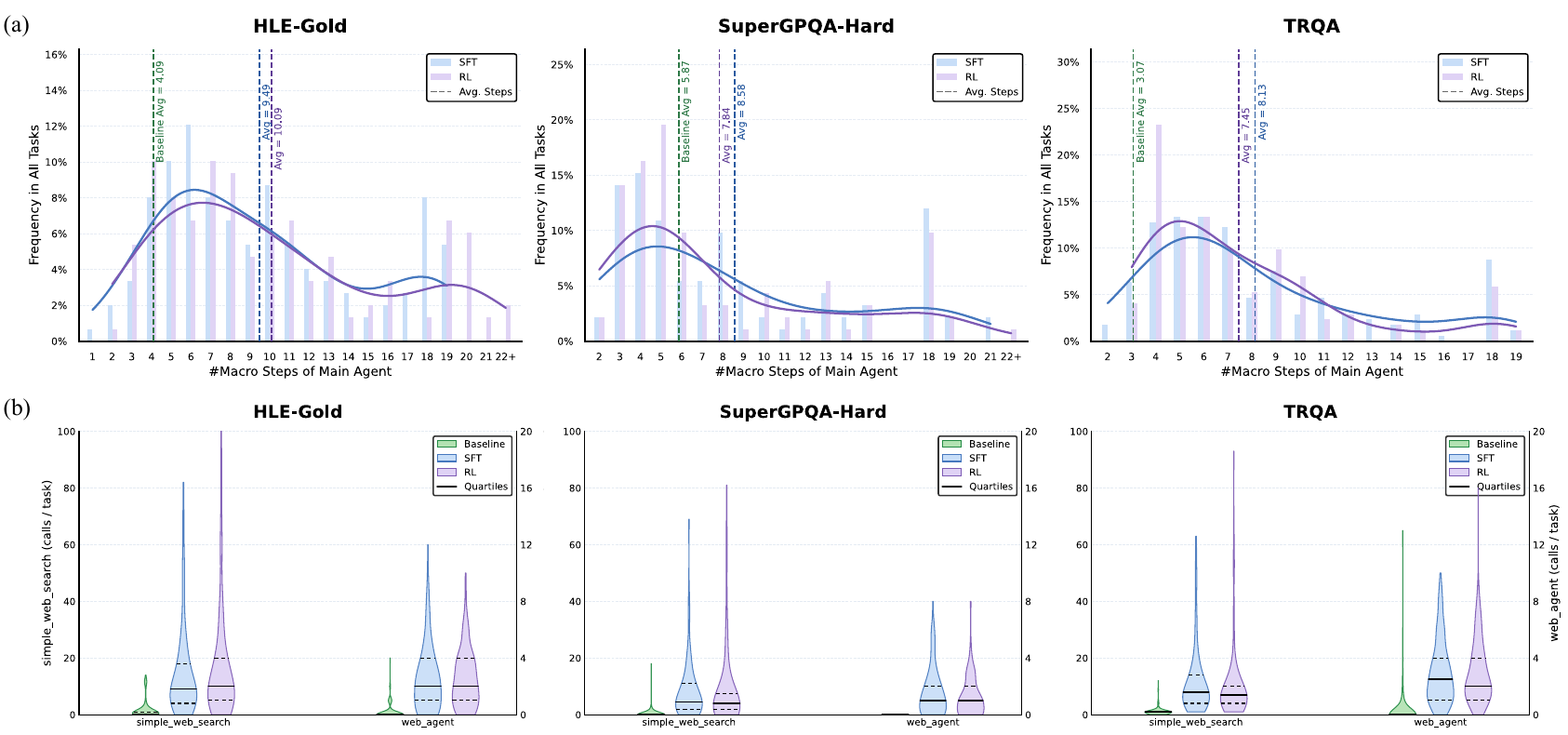}
    \caption{(a) Distribution of trajectory lengths (in macro steps) for SFT and RL checkpoints. (b) Distribution of tool-use frequency for simple web search and web agent.}
    \label{fig:benchmark_traj}
\end{figure*}

\begin{table}[t]
\centering
\small
\begin{tabularx}{\columnwidth}{@{}>{\raggedright\arraybackslash}X cc@{}}
\toprule
\textbf{Training Setting} & \textbf{HLE-Gold} & \textbf{SuperGPQA-Hard} \\
\midrule
Qwen3-8B (baseline) & 8.05 & 22.83 \\
+ Conceptual & 10.74 {\scriptsize(+2.69)} & 25.00 {\scriptsize(+2.17)} \\
+ Computational & 12.08 {\scriptsize(+1.34)} & 28.26 {\scriptsize(+3.26)} \\
+ TRQA \& SciBench & 12.75 {\scriptsize(+0.67)} & 31.52 {\scriptsize(+3.26)} \\
\bottomrule
\end{tabularx}
\caption{Ablation study of training data (cumulative).}

\label{tab:ablation_training_settings}
\vspace{-0.3cm}
\end{table}

\paragraph{Long-Horizon and Tool-Use}
Figure~\ref{fig:benchmark_traj}(a) shows the distribution of trajectory lengths for the baseline, SFT, and RL agents across the three benchmarks. After training, the agent generally produces substantially longer trajectories than the baseline, with trajectory lengths increasing by roughly \(0.3\times\) to \(2.7\times\), indicating a markedly stronger tendency to sustain multi-step exploration and reasoning. A notable pattern emerges when comparing SFT and RL: on the most difficult benchmark, HLE-Gold, the RL checkpoint tends to generate even longer trajectories than the SFT checkpoint, whereas on the other two comparatively easier benchmarks it produces shorter trajectories. This suggests that RL does not simply encourage more steps indiscriminately, but may instead improve the agent's ability to allocate search and reasoning effort more adaptively based on task difficulty.

Figure~\ref{fig:benchmark_traj}(b) provides a complementary view through tool-use statistics, comparing the usage distributions of simple web search and the web-agent tool. Across all three benchmarks, both the SFT and RL checkpoints invoke tools substantially more often than the baseline, indicating that training improves the agent's willingness and ability to rely on external information sources rather than prematurely answering from parametric memory alone. Furthermore, the RL checkpoint typically attains higher maximum tool-use counts than the SFT checkpoint, suggesting a stronger capacity to maintain extended tool-assisted reasoning chains when necessary. Taken together, these findings show that the gains of \textit{\textbf{SciResearcher}}-8B are accompanied by measurable changes in behavior: the model not only answers more accurately, but also exhibits more persistent long-horizon planning and more intensive tool use.

\section{Related Works}
\subsection{Deep Research Agents}
Early progress in deep research agents has been driven by proprietary frontier-model systems~\citep{openai2025deepresearch,citron2024gemini_deep_research,perplexity_deep_research}. In open-source research, recent work mainly focuses on automated data curation for agentic post-training. Existing methods include retrieval-centric pipelines that synthesize supervision from web graphs or entity traversal~\citep{websailor,li2025websailorv2bridgingchasmproprietary,wu2025webwalkerbenchmarkingllmsweb,miromindteam2025mirothinkerpushingperformanceboundaries}, and exploration-centric pipelines that let agents iteratively search and browse to construct harder long-horizon tasks~\citep{tao2025webshaperagenticallydatasynthesizing,webdancer,liu2025webexplorerexploreevolvetraining,wang2025exploreevolvescalingevolved}. These methods improve multi-step information seeking, but their supervision is still largely general-domain, retrieval-oriented, and often targets short factual answers. Other work studies self-evolving agents~\citep{fang2025webevolverenhancingwebagent,li2026verifiedcriticalstepoptimization,wan2026inferencetimescalingverificationselfevolving,hu2025webcotenhancingwebagent}, while recent benchmarks evaluate long-horizon research-style behavior~\citep{wei2025browsecompsimplechallengingbenchmark,futuresearch2025deepresearchbenchevaluating,du2025deepresearchbenchcomprehensivebenchmark}.
\subsection{Agents in Scientific Reasoning}
Prior work improves LLM scientific reasoning by designing specialized agent frameworks and integrating domain-specific knowledge sources. SciMaster~\citep{chai2025scimastergeneralpurposescientificai} uses parallel solution sampling and iterative refinement for general scientific reasoning, Biomni~\citep{huang2025biomni} supports biomedical reasoning with tailored tools and databases, and Eigen-1~\citep{tang2025eigen1adaptivemultiagentrefinement} targets biology and chemistry via retrieval-augmented generation over curated scientific papers. More broadly, LLM agents are increasingly used for scientific research and discovery~\citep{zheng-etal-2025-automation,luo2025llm4srsurveylargelanguage,zheng2026newtonbenchbenchmarkinggeneralizablescientific,ye2026evaluationdrivenscalingscientificdiscovery}, ranging from iterative optimization agents that improve machine-learning models through execution feedback~\citep{jiang2025aideaidrivenexplorationspace,liu2025mlmasteraiforaiintegrationexploration,karpathy2026autoresearch} to autonomous AI scientists that conduct the full research loop from ideation to paper writing~\citep{lu2024aiscientistfullyautomated,weng2025deepscientistadvancingfrontierpushingscientific}.

\section{Conclusion}

\textit{\textbf{SciResearcher}} introduces a fully automated framework for constructing frontier scientific reasoning tasks that integrate conceptual evidence synthesis and computational modeling. By post‑training a Qwen3‑8B agent on the resulting data, we obtain \textit{\textbf{SciResearcher}}-8B, which achieves state‑of‑the‑art performance at its scale and substantially outperforms larger proprietary scientific agents. Our results show that targeted data curation can cultivate adaptive, tool‑intensive reasoning behaviors in compact models. Looking ahead, we see extending this paradigm to additional scientific disciplines and formally characterizing the reasoning taxonomies of frontier science as critical steps toward building fully autonomous scientific agents.

\section*{Limitations}

Despite the strong empirical performance, our work has several limitations.

First, our current study focuses primarily on knowledge-intensive frontier reasoning in biology and chemistry. We do not extensively evaluate domains such as mathematics, physics, materials science, or engineering. These areas may require different forms of reasoning, such as formal proof, symbolic derivation, simulation, or experiment-design capabilities, and it remains an open question how well the proposed data construction paradigm transfers to them.

Second, the scale of our training data is relatively small. Although SciResearcher-8B achieves substantial gains over strong baselines and outperforms several agents built on larger backbones and trained with substantially more data, we have not yet systematically studied data scaling behavior. In particular, the relationship between dataset size, model parameter size, and downstream scientific reasoning performance remains to be characterized.

Third, our experiments are conducted only on an 8B-scale backbone. We believe that the proposed framework is compatible with larger models and may yield stronger results when combined with larger backbones and expanded training data, but this hypothesis has not been empirically validated in this work.

Fourth, the reasoning ontology of our constructed data is limited to two broad categories. Although the free-form nature of conceptual questions can cover some atomic reasoning patterns beyond knowledge inference, it is still insufficient to fully capture the diverse reasoning nature of general science. A promising direction is to curate a fine-grained taxonomy of atomic scientific reasoning skills and synthesize diverse questions compositionally based on this taxonomy~\citep{lee2026drbencheragentidentifyentity}.

Fifth, we instantiate our post-training experiments within the Cognitive Kernel-Pro framework. Although CK-Pro follows a fairly general ReAct-style agent design with planning, tool use, and observation-based decision making, different agent frameworks may expose different action spaces, tool interfaces, memory mechanisms, or planning structures. Therefore, additional experiments are needed to verify the transferability of SciResearcherQA across broader agent architectures.

Finally, although SciResearcher includes multiple postprocessing and verification stages, the generated scientific questions may still contain residual errors, ambiguous wording, incomplete evidence grounding, or unintended reasoning shortcuts. In particular, computational questions are sensitive to model specification, parameter assumptions, and numerical verification. We have not yet conducted a large-scale human expert audit of the dataset, which will be important for further improving reliability and supporting broader scientific use.

\section*{Ethics Statement}
This work develops automated data construction and agent post-training methods for frontier scientific reasoning, which may affect how AI systems are used in scientific research. We use publicly accessible academic sources and automatically synthesized tasks, and we do not intentionally collect private, personal, or sensitive user data. However, scientific agents trained on such data may still produce incorrect, incomplete, or overconfident outputs, especially in high-stakes biomedical or chemical contexts. Therefore, SciResearcher and SciResearcherQA should be viewed as research tools for improving evidence-grounded reasoning rather than substitutes for expert scientific judgment. Any deployment in real-world scientific workflows should involve human expert oversight, careful validation, and attention to potential misuse, including the generation of misleading scientific claims or unsafe biological or chemical guidance. The external benchmarks used in our evaluation, including HLE, SuperGPQA, and TRQA, are publicly available for research use, and our use of them is consistent with their licenses and terms of access. We conducted limited human evaluations through offline private workshops and meetings. Participation was voluntary, feedback was used only in anonymized and aggregated form, and participants were compensated at no less than USD 30 per hour, or the applicable local equivalent.


\bibliography{ref}

\newpage
\appendix

\section{Technical Details}

\subsection{Agent Framework Details}

Our agent framework is built upon Cognitive Kernel-Pro \citep{fang2025cognitivekernelproframeworkdeep}, a two-tier, multi-module hierarchical architecture in which a main agent is responsible for task decomposition, subtask delegation, evidence aggregation, tool invocation, and Python-based action generation, while specialized sub-agents handle grounded interactions with external environments. Web access functionality is supported with Google-Search API and Browserless API. Concretely, we instantiate a web agent for live web navigation and a file agent for local document processing, following the general design of Cognitive Kernel-Pro. Different from the original framework, however, we remove the \texttt{ask\_llm} function from the main agent. This modification prevents the agent from bypassing tool use and producing unsupported shortcut answers directly from the base language model, thereby encouraging explicit evidence collection through interaction with the web and files. To further adapt our agent for frontier scientific reasoning, we refine the system prompt of the main agent to encourage proactive acquisition of domain knowledge and careful verification.

Across all experiments, we optimize only the main agent and use Qwen3-8B (without thinking) as its backbone. The sub-agents remain frozen throughout trajectory sampling, reinforcement learning, and benchmark evaluation.

\begin{table}[H]
\centering
\scriptsize
\setlength{\tabcolsep}{4pt}
\renewcommand{\arraystretch}{1.1}

\begin{tabular}{lll}
\toprule
\textbf{Agent} & \textbf{Action / Tool} & \textbf{Observation} \\
\midrule

\multirow{4}{*}{Main}
& \texttt{web\_agent(task)} & Web-agent output / log \\
& \texttt{file\_agent(task)} & File-agent output / log \\
& \texttt{simple\_web\_search(query)} & Search results \\
& \texttt{stop(answer, summary)} & Final answer \\

\midrule

\multirow{10}{*}{Web}
& \texttt{click(id)} & Web text \& DOM \\
& \texttt{type(id, text)} & Web text \& DOM \\
& \texttt{scroll\_up()} / \texttt{scroll\_down()} & Updated page state \\
& \texttt{wait()} & Updated page state \\
& \texttt{goback()} & Previous page state \\
& \texttt{restart()} & Reset browser state \\
& \texttt{goto(url)} & Web text \& DOM \\
& \texttt{save(remote, local)} & Saved local file / status \\
& \texttt{screenshot(flag, path)} & Screenshot-enabled page view \\
& \texttt{stop(answer, summary)} & Final sub-agent output \\

\midrule

\multirow{5}{*}{File}
& \texttt{load\_file(path)} & File metadata / page index \\
& \texttt{read\_text(path, pages)} & Extracted file text \\
& \texttt{read\_screenshot(path, pages)} & Visual page content \\
& \texttt{search(path, keywords)} & Matched spans / page hits \\
& \texttt{stop(answer, summary)} & Final sub-agent output \\

\bottomrule
\end{tabular}

\caption{Tool and action space of our agent framework.}
\label{tab:agent_tools}
\end{table}

\newpage
\subsection{Baseline Details}

\paragraph{AutoGen} \citep{wu2024autogen} is a multi-agent conversation framework that composes LLMs to collaboratively solve tasks through flexible conversation patterns; in our comparison, it uses GPT-4.1 and follows a standard two-agent scientific QA setup.

\paragraph{SciMaster} \citep{chai2025scimastergeneralpurposescientificai} introduces X-Master, a tool-augmented agent with a four-stage pipeline (Solve, Critic, Rewrite, Summarize) that executes each stage in parallel, and scales further via X-Masters, a stacked multi-agent workflow. It uses code as an interaction language and runs on GPT-4.1.

\paragraph{Biomni} \citep{huang2025biomni} is a biomedical AI agent that first maps the biomedical action space by mining tools, databases, and protocols from publications across 25 subfields, then integrates LLM reasoning with retrieval-augmented planning and code execution to compose complex workflows without templates. It is evaluated with GPT-4.1.

\paragraph{WebSailor} \citep{websailor} is a post-training methodology for Qwen-2.5 web agents (7B, 32B) that synthesizes high-uncertainty QA corpora via knowledge-graph sampling and information obfuscation, followed by rejection-sampling fine-tuning and the DUPO agentic RL algorithm, achieving strong open-source performance on complex web reasoning.

\paragraph{WebThinker} \citep{li2025webthinkerempoweringlargereasoning} extends large reasoning models (e.g., DeepSeek-R1) with a Deep Web Explorer module for dynamic web search and navigation, interleaving reasoning, search, and drafting through a Think-Search-and-Draft strategy. It is optimized with iterative online DPO and available in 7B and 14B versions.

\paragraph{Search-R1} \citep{jin2025searchr1trainingllmsreason} adapts the DeepSeek-R1 RL framework to train LLMs to generate multiple search queries during reasoning with real-time retrieval. Using retrieved token masking for stable RL and outcome-based rewards, Search-R1-7B learns effective multi-turn search interactions.

\newpage
\section{Implementation Details}
\label{app:implementation_details}

Table~\ref{tab:implementation_details} summarizes the key implementation details of our data construction, training, and evaluation pipeline. Unless otherwise specified, all evaluations of our models are conducted within the same Cognitive Kernel-Pro agent framework, using the same tool interfaces, prompt templates, decoding configuration, and inference budget.

\begin{table*}[t]

\centering

\small
\begin{tabular}{p{0.34\linewidth} p{0.58\linewidth}}
\toprule
\textbf{Component} & \textbf{Setting} \\
\midrule
Agent framework & Cognitive Kernel-Pro \\
Model backbone & Qwen3-8B \\
Teacher model for trajectory construction & Claude-Sonnet-4.5 \\
Scientific domains & Biology and chemistry \\
Number of training tasks & 727 \\
Number of step-level training messages & 5,105 \\
Training procedure & Supervised fine-tuning followed by GRPO-style reinforcement learning \\
Reward function & Outcome reward based on final-answer correctness \\
Search tool & Google Search API \\
Browser tool & Browserless API \\
Code execution tool & Python execution environment \\
Maximum agent steps during evaluation & {20} \\
Maximum wall-clock time per question & {1800s} \\
Python execution timeout & {60s} \\
Decoding temperature & {0.2} \\
Top-$p$ & {0.95} \\
Maximum generation length & {4096} \\
Evaluation protocol & Pass@1 and Pass@3 \\
Pass@3 aggregation & Three independent trajectories are sampled for each question; an example is counted as correct if any trajectory produces a correct final answer \\
Answer extraction and judging & LLM Judge (Qwen3-32B) with same instruction in the HLE evaluation, achieving 98.7\% agreement with human judge in HLE-Gold subset. \\
\bottomrule
\end{tabular}
\caption{
Implementation details for SciResearcher data construction, training, and evaluation. 
}
\label{tab:implementation_details}
\end{table*}

For open-source agent baselines, we manually reproduced the reported systems using their official codebases and released model checkpoints. To ensure faithful reproduction, we followed the inference settings specified in the original papers.

\newpage

\section{Data Quality Assessment}
\label{app:data_quality_assessment}

Table~\ref{tab:case_studies} presents one representative question from each pipeline of SciResearcherQA, along with the academic evidence used in its construction. The conceptual example illustrates a multi-hop problem that synthesizes information from three distinct scientific sources, while the computational example demonstrates a scenario-based ODE question built from a published treatment-response model. Together, these examples highlight the multi-source, evidence-grounded, and computationally rich nature of the tasks curated by SciResearcher.

In this section, we present the results of human evaluation and dataset overlap analysis to further assess the generation quality of the SciResearcherQA. We emphasize that SciResearcherQA is intended primarily as scalable training supervision rather than a fully expert-certified scientific benchmark. Human evaluation reveals residual shortcut and evidence-alignment errors, especially in computational tasks, motivating future expert-audited filtering.

\subsection{Human Evaluation}

To assess the quality of the automatically constructed SciResearcherQA examples, we conduct a human evaluation on a randomly sampled subset of the dataset. Specifically, we sample 50 questions from the conceptual QA subset and 50 questions from the computational QA subset. Each example is independently evaluated by three human annotators, all of whom are postgraduate researchers with backgrounds in AI for Science. Annotators judge each example using three binary criteria: \textit{Evidence Entailment}, which measures whether the provided evidence supports the question and the intended reasoning path; \textit{Correct \& Unique Answer}, which measures whether the answer is unambiguous and deductively guaranteed by the question and evidence; and \textit{No Reasoning Shortcuts}, which measures whether the question cannot be correctly answered without using the required evidence. For computational QA, evidence entailment specifically means that the cited evidence paper contains a suitable scientific model for the constructed scenario. We aggregate the three annotations for each example by majority vote and report the resulting pass rates in Table~\ref{tab:human_eval_data_quality}. The overall Fleiss' $\kappa$ across all binary judgments is 0.45, indicating moderate inter-annotator agreement. Overall, SciResearcherQA achieve 89--94\% pass rates across the three metrics, indicating that the constructed examples are generally well grounded and answerable. Nevertheless, the evaluation also reveals several remaining limitations. Although our pipeline includes reasoning-shortcut mitigation, conceptual QA still contains a non-negligible fraction of shortcut cases, suggesting that some examples may require fewer reasoning hops than intended. We consider such cases less harmful for training than factual or answer errors, but they may weaken the intended multi-hop reasoning signal. In addition, the lower evidence-entailment score for computational QA suggests a recurring failure mode: when the generator fails to instantiate the selected equations after multiple attempts, it may fall back to a loosely related formulation that remains topically relevant but is no longer fully entailed by the selected evidence. This highlights the importance of stronger error detection and fallback sensitivity in the generation pipeline.

\begin{table*}[t]

\centering
\small
\begin{tabular}{lccc}
\toprule
\textbf{Subset} 
& \textbf{Evidence Entailment} 
& \textbf{Correct \& Unique Answer} 
& \textbf{No Reasoning Shortcuts} \\
\midrule
Conceptual QA 

& {98\%} 
& {94\%} 
& {82\%} \\
Computational QA 

& {86\%} 
& {94\%} 
& {96\%} \\
\midrule
Overall 
& 92\%
& 94\%
& 89\% \\
\bottomrule
\end{tabular}
\caption{
Human evaluation of SciResearcherQA data quality. Each example is independently judged by three postgraduate AI4Science annotators using binary criteria, and the reported result is based on majority vote. Note: In the context of computational questions, ``Evidence Entailment'' measures whether the evidence paper contains the suitable scientific model for the given scenario.
}
\label{tab:human_eval_data_quality}
\end{table*}

\subsection{Dataset Overlap Analysis}
\label{app:overlap}

To assess whether the observed benchmark improvements are likely to reflect generalization rather than direct dataset overlap, we conduct a multi-dimensional overlap analysis between our synthesized training corpora---\textbf{SciResearcherQA-Concept} and \textbf{SciResearcherQA-Compute}---and the three evaluation benchmarks.
We examine four complementary signals: (i) near-duplicate detection at multiple similarity thresholds, (ii) semantic and lexical similarity using combined embeddings, (iii) biomedical entity overlap, and (iv) domain distribution alignment.
While such analyses cannot rule out all possible forms of contamination, especially overlap through external web sources or model pretraining corpora, they provide a useful check for direct question-level overlap between our constructed data and the evaluation sets.

\paragraph{Near-duplicate detection.}
We first perform strict near-duplicate detection across all dataset pairs.
For each pair of datasets, we compute a combined cosine similarity score between all question pairs, using a weighted combination of sentence embeddings from all-MiniLM-L6-v2 \citep{wang2020minilmdeepselfattentiondistillation} and TF-IDF features, with weights of 70\% and 30\%, respectively.
We then count the number of question pairs whose similarity exceeds thresholds of 0.80, 0.85, and 0.90.
As shown in Table~\ref{tab:near_dup}, we do not detect any question pairs above these thresholds across the analyzed dataset pairs.
This suggests that the synthesized corpora do not contain verbatim or near-verbatim duplicates of the benchmark questions under our similarity metric.

\begin{table*}[h]
\centering
\small
\begin{tabular}{lccc}
\toprule
\textbf{Dataset Pair} & \textbf{\(\geq\)0.80} & \textbf{\(\geq\)0.85} & \textbf{\(\geq\)0.90} \\
\midrule
Conceptual \(\leftrightarrow\) Computational & 0 & 0 & 0 \\
Conceptual \(\leftrightarrow\) HLE & 0 & 0 & 0 \\
Conceptual \(\leftrightarrow\) TRQA & 0 & 0 & 0 \\
Conceptual \(\leftrightarrow\) SuperGPQA & 0 & 0 & 0 \\
Computational \(\leftrightarrow\) HLE & 0 & 0 & 0 \\
Computational \(\leftrightarrow\) TRQA & 0 & 0 & 0 \\
Computational \(\leftrightarrow\) SuperGPQA & 0 & 0 & 0 \\
HLE \(\leftrightarrow\) TRQA & 0 & 0 & 0 \\
HLE \(\leftrightarrow\) SuperGPQA & 0 & 0 & 0 \\
TRQA \(\leftrightarrow\) SuperGPQA & 0 & 0 & 0 \\
\bottomrule
\end{tabular}
\caption{Near-duplicate pair counts at varying similarity thresholds.}
\label{tab:near_dup}
\end{table*}

\paragraph{Max-neighbor similarity.}
We next compute, for each question in a target dataset, its maximum similarity to any question in a source dataset.
This max-neighbor statistic captures the closest question-level match across two datasets and therefore provides a stricter view of potential memorization risk than mean pairwise similarity alone.
As reported in Table~\ref{tab:max_neighbor}, the mean max-neighbor similarity between SciResearcherQA-Concept and the evaluation benchmarks is 0.272, close to the benchmark--benchmark average of 0.259.
The highest average max-neighbor value among the analyzed Conceptual--benchmark pairs is 0.348 for Conceptual \(\rightarrow\) TRQA, which remains well below the near-duplicate thresholds used above.
These results suggest that the Conceptual subset is not unusually close to the evaluation benchmarks relative to the similarity observed among the benchmarks themselves.

\begin{table*}[h]
\centering
\small
\begin{tabular}{lcc}
\toprule
\textbf{Metric} & \textbf{Conceptual \(\leftrightarrow\) Benchmarks} & \textbf{Benchmark \(\leftrightarrow\) Benchmark} \\
\midrule
\rowcolor{TableHeader}
\textit{Mean max-neighbor sim.} & 0.272 & 0.259 \\
\textit{Mean pairwise sim.} & 0.096 & 0.080 \\
\rowcolor{TableHeader}
\textit{Entity Jaccard overlap} & 0.079 & 0.051 \\
\textit{Domain cosine sim.} & 0.436 & 0.677 \\
\bottomrule
\end{tabular}
\caption{Mean nearest-neighbor, pairwise, entity-overlap, and domain-distribution statistics.}
\label{tab:max_neighbor}
\end{table*}

\paragraph{Mean pairwise and entity-level overlap.}
The mean pairwise combined similarity between SciResearcherQA-Concept and the evaluation benchmarks is 0.096, only slightly higher than the 0.080 observed among the benchmarks themselves, as shown in Table~\ref{tab:max_neighbor}.
This modest increase is expected in biomedical and chemical QA settings, where datasets often share domain-specific terminology such as gene names, pathways, diseases, drugs, and experimental techniques.
We therefore interpret this signal as reflecting shared scientific vocabulary rather than direct structural duplication.
Consistent with this interpretation, biomedical entity overlap remains low in absolute terms: the gazetteer-based Jaccard similarity over genes/proteins, diseases, drugs, and cell types is below 0.08 for the Conceptual--benchmark comparison.

\paragraph{Domain distribution.}
We also compare the topical coverage of the datasets.
Domain classification is performed by keyword matching into eleven biomedical sub-domains, including Genetics, Biochemistry, Pharmacology, Oncology, Immunology, and Computational Modeling.
For each dataset, we construct a normalized frequency vector over these sub-domains and compute cosine similarity between vectors.
As shown in Table~\ref{tab:max_neighbor}, the domain-distribution cosine similarity is 0.436 for SciResearcherQA-Concept versus the benchmarks, compared with 0.677 among the benchmarks themselves.
This indicates that SciResearcherQA-Concept is not more closely aligned with the benchmark domain distribution than the benchmarks are with each other.
Instead, it appears to cover a broader and somewhat more diffuse set of sub-domains.

\paragraph{Intra-dataset diversity.}
Table~\ref{tab:intra} reports intra-dataset mean pairwise similarity, where lower values indicate greater internal diversity.
SciResearcherQA-Concept has a self-similarity of 0.131, which is comparable to SuperGPQA (0.164) and lower than TRQA (0.198).
HLE has the lowest self-similarity (0.078), reflecting its broad and heterogeneous scope.
SciResearcherQA-Compute has higher self-similarity (0.240), which is expected because it is intentionally focused on computational-modeling questions.
Overall, these results suggest that the synthesized data is not narrowly concentrated around a small set of repeated templates or highly similar questions.

\begin{table*}[h]
\centering
\small
\begin{tabular}{lc}
\toprule
\textbf{Dataset} & \textbf{Self-Similarity} \\
\midrule
HLE & 0.078 \\
Conceptual & 0.131 \\
SuperGPQA & 0.164 \\
TRQA & 0.198 \\
Computational & 0.240 \\
\bottomrule
\end{tabular}
\caption{Intra-dataset diversity measured by mean pairwise similarity. Lower values indicate greater diversity.}
\label{tab:intra}
\end{table*}

\begin{figure*}[t]
\centering
\includegraphics[width=\textwidth]{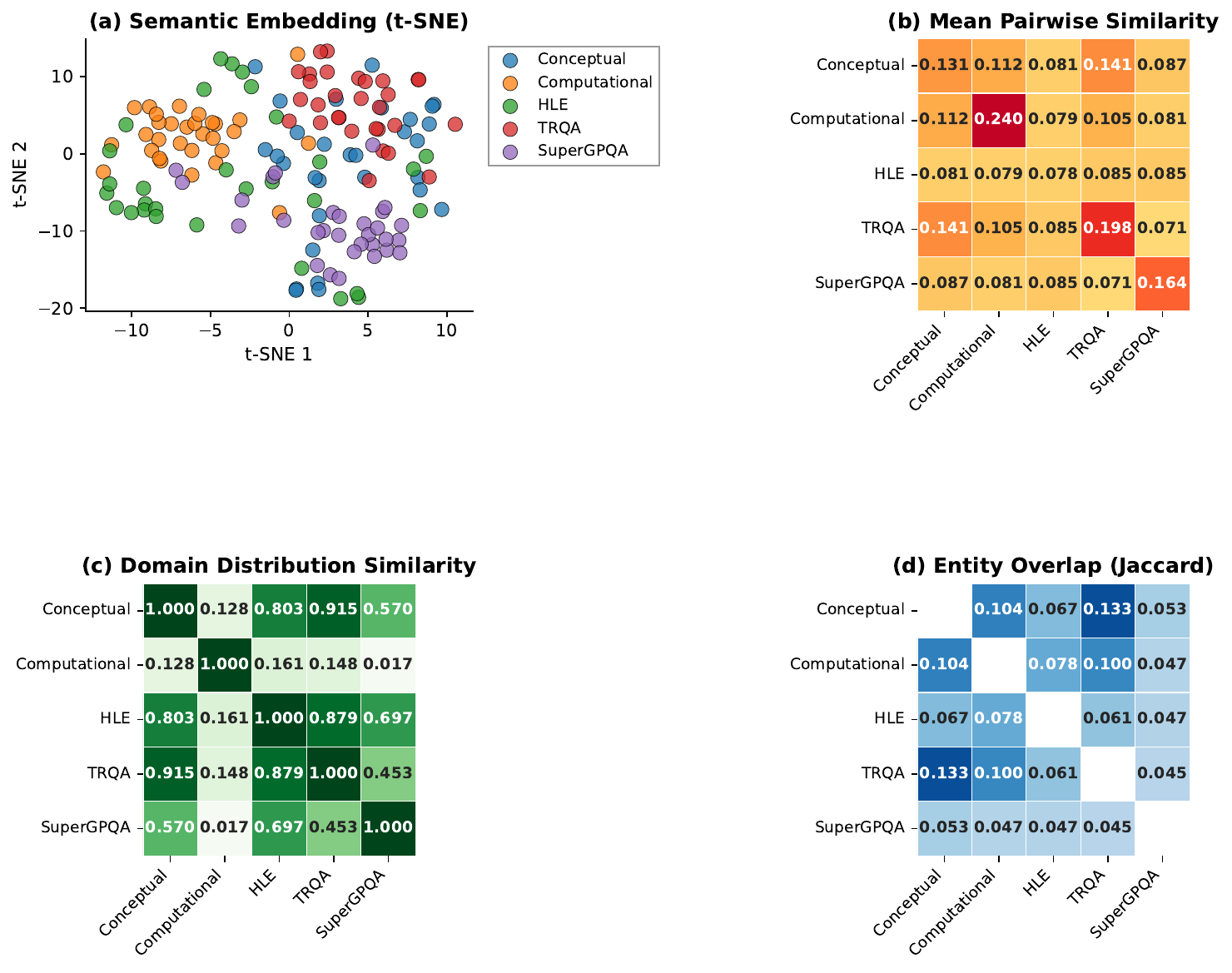}
\vspace{1.6cm}

\includegraphics[width=\textwidth]{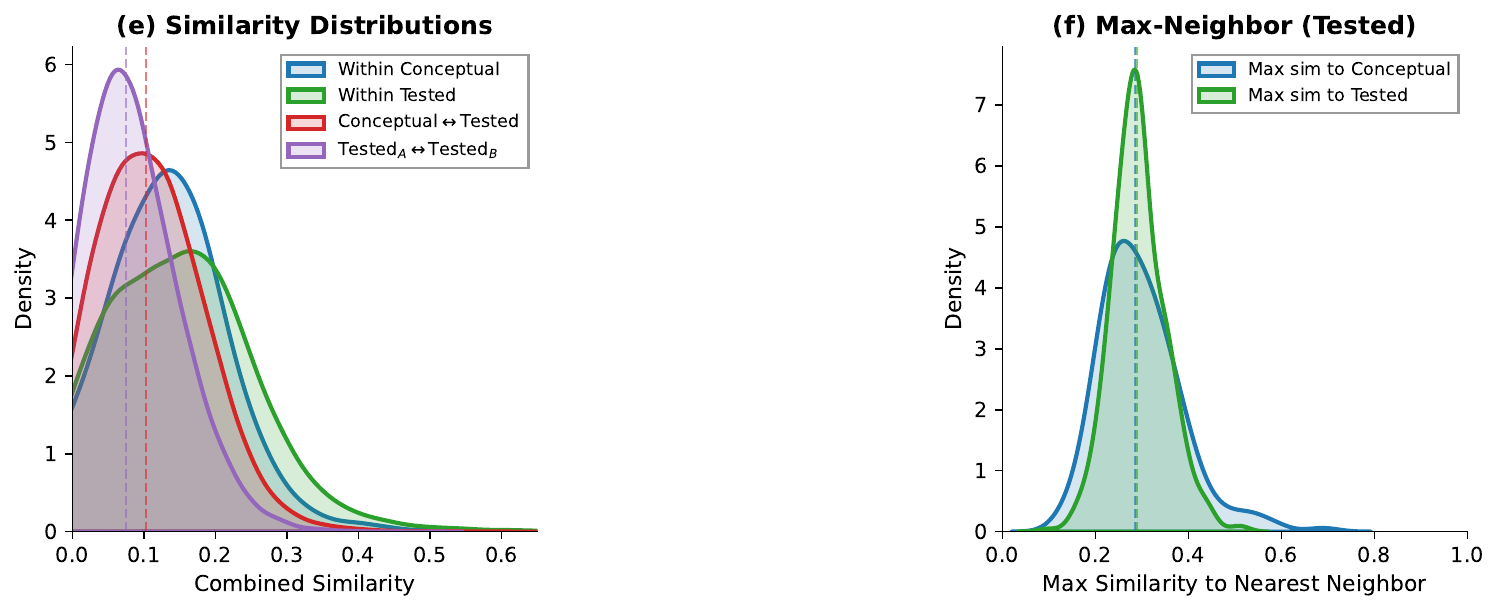}
\caption{Dataset overlap analysis. 
(a) t-SNE projection of question embeddings, using 30 sampled questions per dataset. 
(b) Mean pairwise combined similarity across datasets. 
(c) Domain-distribution cosine similarity, where higher values indicate more similar topic coverage. 
(d) Dataset-level biomedical entity overlap measured by Jaccard similarity. 
(e) Kernel density estimates of pairwise similarity distributions. 
(f) Max-neighbor similarity distributions for benchmark questions, comparing SciResearcherQA-Concept with other benchmark datasets. 
Dashed vertical lines indicate medians.}
\label{fig:dist}
\end{figure*}

\paragraph{Visual summary.}
Figure~\ref{fig:dist} provides a visual summary of the semantic and distributional relationships between the synthesized corpora and the evaluation benchmarks.
It includes a t-SNE projection of question embeddings, heatmaps of mean pairwise similarity, domain-distribution similarity, and entity overlap, as well as kernel density estimates of pairwise similarity and max-neighbor similarity distributions.
These visualizations are consistent with the quantitative results above: the synthesized corpora occupy a broad region of the biomedical QA space while showing limited direct overlap with the evaluated benchmarks.

\vspace{0.5em}
\noindent\textbf{Takeaway.}
Across several complementary measures, we find little evidence of direct question-level overlap between SciResearcherQA and the evaluation benchmarks.
In particular, we observe no near-duplicate question pairs above the tested thresholds, low nearest-neighbor similarities, low entity-level overlap, and domain distributions that are not unusually aligned with the benchmarks.
These findings support the interpretation that the performance gains from SciResearcher training are unlikely to be driven by direct benchmark memorization, although they do not eliminate broader contamination risks from external sources or pretrained model exposure.

\section{Prompt Templates}
Tables~\ref{tab:prompt_seed2question}--\ref{tab:prompt_eval_urls} document the complete prompt templates employed in the conceptual and computational task curation pipelines. Each template is presented in its original form, covering task identity, input/output format, selection criteria, and workflow instructions. Together, they specify the agentic procedures described in Section~2.


{\renewcommand{\arraystretch}{1.1}
\begin{table*}[t]
\centering

\scriptsize

\resizebox{\textwidth}{!}{
\begin{tabular}{@{}p{0.10\textwidth}
                 p{0.48\textwidth}
                 p{0.38\textwidth}@{}}
\toprule
\textbf{Type} & \textbf{Example Question} & \textbf{Supporting Evidence} \\
\midrule

\textbf{Conceptual}
&
\hlone{A vitamin whose deficiency causes concurrent acute encephalopathy and
hepatic steatosis when induced by a specific antagonist}
binds to \hltwo{a protein with the following properties}:
\hltwo{(1) contains only 5 N-terminal domain disulfide bonds, unlike other family
members with 6;}
\hltwo{(2) exhibits more than 100-fold selectivity for IGF-II over IGF-I;}
and
\hltwo{(3) has C-terminal residues Ser180, Ser181, and Gln182 determining its
IGF-II specificity.}
\hlthree{What is the binding site and interaction energy?}

\smallskip
\textit{Answer:} ASN-29 with binding energy of -4.3 kcal/mol
&
{\scriptsize
\hlone{\textit{Source 1:}} Vitamins and Minerals for Energy, Fatigue and
Cognition: A Narrative Review of the Biochemical and Clinical Evidence
(\textit{Nutrients}, 2020).

\smallskip
\hltwo{\textit{Source 2:}} Insulin-Like Growth Factor Binding Proteins: A
Structural Perspective (\textit{Front.\ Endocrinol.}, 2012).

\smallskip
\hlthree{\textit{Source 3:}} Pantothenic acid ameliorates hepatic fibrosis by
targeting IGFBP6 to regulate the TGF-$\beta$/SMADs pathway
(\textit{Commun.\ Biol.}, 2025).}
\\

\addlinespace[2pt]
\midrule

\textbf{Computational}
&
A 42-year-old patient with a WHO Grade~II, 1p/19q-codeleted
oligodendroglioma is treated with temozolomide (TMZ) chemotherapy. Let
\(P(t)\) denote the volume of proliferative tumor cells, \(D(t)\) the
volume of lethally damaged cells, and \(C(t)\) the intratumoral TMZ
concentration, with total tumor volume \(V(t)=P(t)+D(t)\).
\hlone{The system follows the standard compartmental treatment-response equations with
logistic growth limitation and two TMZ-induced effects.}
The patient completes three full treatment cycles (84 days total), with dosing on days
1,\,2,\,3,\,4,\,5,\,29,\,30,\,31,\,32,\,33,\,57,\,58,\,59,\,60,\,61.
What is the total tumor volume \(V(84)\) in \(\mathrm{cm}^{3}\),
immediately after completion of the third cycle?

\smallskip
\textit{Answer:} 62.77 cm\(^{3}\)
&
{\scriptsize
\hlone{\textit{Source:}} Computational design of improved standardized
chemotherapy protocols for grade~II oligodendrogliomas
(\textit{PLOS Comput.\ Biol.}, 2019).

\smallskip
\textit{Model equations} (ODE system):
\vspace{-0.07cm}
{\setlength{\abovedisplayskip}{3pt}
 \setlength{\belowdisplayskip}{3pt}
 \setlength{\abovedisplayshortskip}{3pt}
 \setlength{\belowdisplayshortskip}{3pt}
 \setlength{\jot}{2pt}
\[
\begin{aligned}
\frac{dP}{dt} &= \rho P\!\left(1-\frac{P+D}{K}\right)
                 - \alpha_{1}PC - \alpha_{2}PC,\\
\frac{dD}{dt} &= \alpha_{1}PC
                 - \frac{\rho}{\kappa}D\!\left(1-\frac{P+D}{K}\right),\\
\frac{dC}{dt} &= -\lambda C.
\end{aligned}
\]
}
}
\vspace{-0.07cm}
\\
\bottomrule
\end{tabular}
}
\caption{Examples of questions and supporting evidence for both question types in SciResearcherQA. Supporting evidence and its corresponding question content are highlighted using the same color.}
\label{tab:case_studies}
\end{table*}
}
{\renewcommand{\arraystretch}{1.1}
\begin{table*}[t]
\centering

\scriptsize
\resizebox{\textwidth}{!}{
\begin{tabular}{@{}p{0.10\textwidth}
                 p{0.30\textwidth}
                 p{0.58\textwidth}@{}}
\toprule
\textbf{Type} & \textbf{Example Question} & \textbf{Supporting Evidence and Diagnosis} \\
\midrule

\textbf{Conceptual}
&
\hlone{A cleavable biotin linker used in peptide-centric chemoproteomics
demonstrates approximately 2-fold higher reproducible cysteine
identifications compared to azobenzene-based linkers, leaves a
+181.1~Da residual mass after cleavage, and exhibits no artifactual
modifications.}
\hltwo{What are the optimal cleavage conditions for this linker?}

\smallskip
\textit{Answer:} 10\% formic acid, 2 hours, room temperature.
{\color{red}\textbf{\textit{(Insufficient Obfuscation)}}}
&
{\scriptsize
\hlone{\textit{Hop 1 --- Linker identification.}}
Evaluation and Optimization of Chemically-Cleavable Linkers for
Quantitative Mapping of Small Molecule-Protein Interactomes
(\textit{ACS Chem. Biol.}, 2019).

\smallskip
\textit{Evidence:} In K562/IAAyne peptide enrichment experiments,
DADPS yields approximately 2$\times$ more reproducible cysteine
identifications than AZO, leaves a +181.1~Da hydroxyl residual on the
peptide after cleavage, and shows no artifactual sulfation.
{\color{red}\textbf{\textit{(Potential Shortcut)}}}

\smallskip
\textit{Diagnosis:} The example is evidence-supported, but the
combination of ``+181.1~Da residual mass,'' improved cysteine
identification over AZO, and absence of artifactual sulfation
\textbf{forms a highly distinctive fingerprint for DADPS}. Thus, a domain
expert may infer the linker identity from the first-hop description alone,
without using the intended evidence. This can reduce the task from explicit
two-hop search to knowledge-based entity inference followed by one-hop
search.

\smallskip
\hltwo{\textit{Hop 2 --- Cleavage condition lookup.}}
Benchmarking Cleavable Biotin Tags for Peptide-Centric Chemoproteomics
(\textit{J. Proteome Res.}, 2022).

\smallskip
\textit{Evidence:} Once the linker is identified as DADPS, the cited
benchmark reports the optimal cleavage condition as
\textbf{10\% formic acid for 2~h at room temperature}.
{\color{red}\textbf{\textit{(Valid and Decisive)}}}
}
\\

\addlinespace[2pt]
\midrule

\textbf{Computational}
&
\hlone{A researcher studies a simplified two-pathway carbon metabolism
network using [U-$^{14}$C]-glucose tracer to estimate pathway flux
partitioning.} The network contains two parallel pathways from
glucose-6-phosphate (G6P) to pyruvate (Pyr): \textit{<pathway description
and parameters omitted>}.
\hltwo{The relationships linking the measured isotopologue fractions to
the pathway fluxes and efficiencies are governed by the
\textbf{conserved-moiety fluxomics framework for isotopic tracer metabolic
flux analysis}.}
Given the measured M+3 fraction, calculate $v_A$. Then use the measured
M+2 fraction and the calculated $v_A$ to solve for $v_B$. Finally, compute
the fractional flux partitioning ratio $v_B/v_A$.

\smallskip
\textit{Answer:} 0.5833.
{\color{red}\textbf{\textit{(Evidence--Scenario Mismatch)}}}
&
{\scriptsize
\hlone{\textit{Selected evidence.}}
Conserved Moiety Fluxomics
(2024).

\smallskip
\textit{What the paper provides:} a general computational framework for
isotopic tracer metabolic flux analysis based on conserved-moiety
transitions and network-scale optimization.

\smallskip
\textit{What the question uses:} a simplified two-pathway toy network with
hand-specified labeling efficiencies and pyruvate isotopologue balance
relations.

\smallskip
{\color{red}\textbf{\textit{Issue: weak alignment between the cited model
and the instantiated calculation.}}}

\smallskip
\textit{Diagnosis:} The cited framework is relevant to the broad topic of
isotopic tracer flux analysis, but it does not directly provide the simple
two-pathway balance equations needed for this particular calculation.
Intuitively, the numerical answer follows from the hand-specified
efficiencies and the observed M+2/M+3 fractions, rather than from a
distinctive component of the conserved-moiety fluxomics framework.
Therefore, the evidence mainly supports the general modeling context,
whereas the instantiated computational closure is introduced by the question itself.
}
\\

\bottomrule
\end{tabular}
}
\caption{
Case study of generated examples with residual quality issues. The conceptual
example is factually grounded but contains a reasoning shortcut because the
first-hop property bundle strongly identifies the linker, reducing the intended
multi-hop reasoning requirement. The computational example illustrates weaker
alignment between the cited modeling framework and the instantiated numerical
calculation.
}
\label{tab:case_study_residual_quality_issues}
\end{table*}
}


\begin{table*}[!ht]
\centering
\begin{tabular}{@{}p{\textwidth}@{}}
\toprule
\begin{minipage}{\textwidth}
\scriptsize
\begin{verbatim}
## Core Identity
<Role: Frontier Scientific Question Curation Agent. Your goal is to generate a base conceptual scientific reasoning question
from a given seed entity, grounded in verifiable academic evidence. The target domains include frontier scientific areas such
as biology, chemistry, biomedicine, and related interdisciplinary fields.>

## Input Format
<Seed entity, together with its domain and ontology information; illustrative examples.>

## Question Curation Requirements
<Metric Definition>
<The generated question should:
1. Include the seed entity or be directly grounded in it.
2. Be concise but scientifically meaningful.
3. Be answerable from a single authoritative academic source at this stage.
4. Prefer multiple-choice format with plausible confounders, while allowing short-answer format when more appropriate.
5. Avoid shortcuts that can be solved by trivia, superficial keyword matching, or generic web search without reading the
   academic evidence.
6. Be suitable as the semantic backbone for later anchor-based augmentation.>

## Pre-Action Protocol: Plan Before Searching
<Metric Definition>
<Before browsing, understand the seed entity and its scientific context. Plan 3--5 diverse search queries that target
academic sources such as peer-reviewed papers, domain databases, preprints, and reputable scientific venues. Assess source
quality based on relevance, authority, evidence specificity, and whether the source supports a nontrivial scientific claim.>

## Question Curation Strategies
<Metric Definition>
<Scoring Description and Examples for each numbered strategy:>
1. Meticulousness and persistence in finding high-quality academic evidence.
2. Task decomposition: search -> evidence extraction -> question generation -> verification.
3. Adaptive error handling and reuse of progress state when searches fail or evidence is insufficient.
4. Multi-query scout search and URL selection based on relevance, venue quality, source diversity, and scientific specificity.
5. Use of the url2evidence sub-agent to access selected academic sources, extract key supporting evidence, and distinguish
   stand-alone scientific facts from study-specific artifacts.
6. Evidence quality checks, including source authority, evidence-answer entailment, and avoidance of unsupported assumptions.
7. Question formulation with plausible, unbiased, and challenging confounders for MCQs; clear expected answer for short-answer
   questions; and final quality checks.
8. Multi-tool coordination following the typical workflow:
   scout search -> source selection -> url2evidence -> question generation -> verification.

## Output Format
The final output MUST be a JSON object with the following structure:
'''json
{
  "question": "The question text containing or directly grounded in the seed entity",
  "answer": "The correct answer content, not a letter label",
  "question_type": "mcq",
  "confounders": ["confounder1", "confounder2", "confounder3"],
  "evidence": {
    "url": "https://...",
    "paper_title": "Title of the paper or academic source",
    "evidence_paragraph": "The exact paragraph or quote that supports the answer",
    "context": "Additional background information that explains the broader context of the evidence or clarifies details needed
    to understand it"
  }
}
'''

**Important Notes on Output Format:**
- 'answer': Provide the actual answer content, not an option letter.
- 'question_type': Use lowercase "mcq" or "short_answer".
- 'confounders': For MCQs, provide 3 or more challenging wrong answers.
- 'evidence.url': Must be a real, verified URL.
- The entire output must be valid JSON and contain no extraneous commentary.
\end{verbatim}
\end{minipage}
\\
\bottomrule
\end{tabular}
\caption{Prompt template for conceptual task curation: \texttt{seed2question}.}
\label{tab:prompt_seed2question}

\end{table*}

\begin{table*}[!ht]
\centering
\label{tab:prompt_anchor_extraction}
\begin{tabular}{@{}p{\textwidth}@{}}
\toprule
\begin{minipage}{\textwidth}
\scriptsize
\begin{verbatim}
You are an expert at analyzing scientific reasoning questions.

Your task is to identify the single most critical "anchor entity" in the question body. This anchor will be used for anchor-based
question augmentation: a new question will later be generated whose answer is exactly this anchor entity, and that new question
will be fused back into the original question.

## Definition of Anchor Entity

An anchor entity is a SPECIFIC scientific term that:
1. **Domain-specific**: It is a concrete scientific entity, such as a gene, protein, pathway, compound, species, technique,
   disease, mutation, phenotype, material, model, or other scientific concept.
2. **Question-body only**: It appears in the question stem but does NOT appear in the correct answer or any confounder.
3. **Decisive**: The question becomes substantially harder or unanswerable if this entity is masked or removed.
4. **Specific and concrete**: It is sufficiently specific to support further evidence-grounded browsing and question generation.

## Your Task

Given the question, correct answer(s), and confounders below, you must:
1. Identify candidate anchor entities in the question body.
2. Verify that each candidate does NOT appear in the correct answer or any confounder.
3. Evaluate whether each candidate is decisive for deriving the final answer.
4. Select the most decisive, specific, and concrete entity.
5. If no valid anchor exists, return an empty string.

## Selection Criteria (in priority order)

1. Prefer the MOST SPECIFIC entity, e.g., "AXL" over "receptor tyrosine kinase".
2. Prefer entities that constrain the answer, such that removing them makes multiple answers plausible.
3. Prefer named entities, such as gene, protein, compound, disease, pathway, or model names, over generic scientific terms.
4. Prefer entities that are decoupled from the surface form of the answer options.
5. If multiple candidates exist, choose the one most central to the scientific claim.

## Output Format

Return ONLY valid JSON:
{
  "candidates": [
    {
      "entity": "...",
      "in_question": true,
      "in_options": false,
      "is_decisive": true
    }
  ],
  "anchor_entity": "<the single valid anchor entity string, or empty string if none>",
  "entity_type": "<type: gene|protein|pathway|compound|technique|disease|other>",
  "reasoning": "<brief explanation of selection and validation>"
}

## Examples
<Illustrative worked examples omitted here for brevity.>

---

## Question
{question}

## Correct Answer(s)
{answer}

## Confounders (wrong options)
{confounders}

Analyze candidates, verify constraints, and return the valid anchor entity.
\end{verbatim}
\end{minipage}
\\
\bottomrule
\end{tabular}
\caption{Prompt template for conceptual task curation: \texttt{anchor\_extraction}.}

\end{table*}

\begin{table*}[!ht]
\centering
\label{tab:prompt_seed2equation}
\begin{tabular}{@{}p{\textwidth}@{}}
\toprule
\begin{minipage}{\textwidth}
\scriptsize
\begin{verbatim}
## Core Identity
<Frontier Scientific Model Discovery Agent. Your goal is to identify an advanced computational or numerical scientific model
associated with a given seed entity, extract its governing equations and application constraints from academic sources, and
prepare the model specification for downstream computational question generation.>

## CARDINAL RULE: Precision and Groundedness Above All
<Metric Definition>
<All extracted model details must be traceable to the selected academic source. Prefer precise, reproducible mathematical
definitions over vague model descriptions. Abandon a candidate model if its equations, parameters, scenario, or source cannot
be verified.>

## Input Format
<Seed entity, together with its domain and ontology information; illustrative examples.>

## Task Overview -- Three-Level Evidence Selection and Model Extraction

### Level 1: Scout Search
Perform multiple scout searches to identify promising academic sources associated with the seed entity. Use diverse queries that
target computational models, numerical simulations, mechanistic equations, kinetic models, ODE/PDE systems, statistical models,
or other quantitative formulations.

### Level 2: URL Evaluation with eval_urls
Use the eval_urls tool to assess selected sources. Prioritize sources according to:
1. Model exclusiveness
2. Search identifiability
3. Computational complexity
4. LLM unfamiliarity

Also consider URL validity and whether the source clearly contains a usable computational or numerical model.

### Level 3: Detailed Model Extraction with url2evidence
Use the url2evidence sub-agent to conduct a deep dive into the final selected source or sources. Extract the complete model
specification, including:
1. Model name and scientific purpose.
2. Governing equations.
3. Variable definitions.
4. Parameter definitions and units.
5. Applicable scenario and constraints.
6. Any assumptions required for correct model use.

## Model Selection Criteria
Select a model that satisfies as many of the following criteria as possible:
1. The model supports calculable numerical outputs.
2. The model is described in a real, citable academic source.
3. The equations are nontrivial and not merely standard textbook formulas.
4. The computation requires meaningful model instantiation or numerical solving.
5. The model can support a realistic scenario-based scientific question.
6. The source is relatively recent, niche, or unlikely to be memorized by LLMs.
7. The model is clearly associated with the seed entity or its scientific domain.

## What Counts as a Frontier Numerical Model?
<A model with explicit mathematical structure, such as governing equations, ODE/PDE systems, kinetic models, dose-response
models, mechanistic simulations, quantitative biological or chemical models, or other computational formulations that can be
instantiated to produce a numerical answer.>

## What Does NOT Count
<Do not select models that are only conceptual diagrams, purely descriptive frameworks, simple textbook equations, standard
unit conversions, or models whose parameters and equations cannot be verified from the source.>

## Output Format
The final output MUST be a JSON object with the following structure:
'''json
{
  "seed_entity": "<the seed entity>",
  "selected_model": {
    "title": "<paper title>",
    "url": "<URL of the paper or academic source where the model is described>",
    "description": "<brief description and its applicable scenario>",
    "equations": "<explicit named equations; see Equation Naming Convention below>",
    "variables": "<definitions of variables used in the equations>",
    "parameters": "<definitions, values if available, and units of parameters>",
    "assumptions": "<model assumptions and application constraints>"
  }
}
'''

**Critical Notes on Output:**
- 'selected_model.url' MUST be a real, verified URL, not a fabricated one.
- 'equations' MUST contain explicit mathematical definitions, not vague descriptions.
- Include all variables, parameters, units, assumptions, and constraints needed for downstream question generation.
- The entire output must be parsable as JSON and contain no extraneous text or commentary.

## Equation Naming Convention
<Use bracketed 5--20 word names for each equation. Each name should describe the scientific role of the equation, e.g.,
[Logistic tumor growth with drug-induced cell damage]. Avoid generic names such as [Equation 1] or [Main formula].>
\end{verbatim}
\end{minipage}
\\
\bottomrule
\end{tabular}
\caption{Prompt template for computational task curation: \texttt{seed2equation}.}

\end{table*}

\begin{table*}[!ht]
\centering
\begin{tabular}{@{}p{\textwidth}@{}}
\toprule
\begin{minipage}{\textwidth}
\scriptsize
\begin{verbatim}
You are an expert scientific model evaluator. You are given the text content of a scientific article or paper. Your task is to
evaluate whether this article contains a computational or numerical model suitable for generating a benchmark question that tests
an AI agent's ability to:
1. Search for and identify the relevant model.
2. Extract the model equations and constraints from the paper.
3. Instantiate the model in a concrete scientific scenario.
4. Write and execute a Python solver to compute a numerical answer.

First, perform preliminary validity checks. Then evaluate the article according to the four core metrics used for computational
task curation.

## Preliminary Check 1: URL Validity
<Metric Definition>
<Determine whether the URL corresponds to a real and accessible academic source, such as a peer-reviewed paper, preprint,
official proceedings page, or reputable scientific database entry.>

## Preliminary Check 2: Model Presence
<Metric Definition>
<Determine whether the article contains an explicit computational or numerical model with equations, variables, parameters, or
algorithmic procedures that can be used to compute a numerical answer.>

## Core Metric 1: Model Exclusiveness (0-10) -- CRITICAL
<Metric Definition>
<Score how specific and source-dependent the model is. High-scoring models have equations, assumptions, or parameterizations
that are distinctive to this paper or research line, rather than generic textbook formulas.>

## Core Metric 2: Search Identifiability (0-10)
<Metric Definition>
<Score whether an agent could plausibly find the source through web search from the seed entity, scientific context, or
model-related clues. A good source should be searchable but not trivially obvious.>

## Core Metric 3: Computational Complexity (0-10)
<Metric Definition>
<Score whether the model requires nontrivial quantitative reasoning, numerical simulation, equation solving, or careful
parameter instantiation. Avoid models that require only simple arithmetic or direct lookup.>

## Core Metric 4: LLM Unfamiliarity (0-10)
<Metric Definition>
<Score how unlikely the model and its exact equations are to be memorized by a general-purpose LLM. Niche, recent, specialized,
or paper-specific models should receive higher scores.>

## Output Format -- strict JSON
'''json
{
  "is_valid_url": true,
  "includes_model": true,
  "model_exclusiveness": 8,
  "search_identifiability": 7,
  "computational_complexity": 8,
  "llm_unfamiliarity": 9,
  "model_name": "<name of the model if identifiable, else 'N/A'>",
  "model_summary": "<1-2 sentence summary of what the model computes>",
  "rationale": "<brief rationale for your validity checks and scores>"
}
'''

Return ONLY the JSON object, with no commentary before or after.
\end{verbatim}
\end{minipage}
\\
\bottomrule
\end{tabular}
\caption{Prompt template for computational task curation: \texttt{eval\_urls}.}
\label{tab:prompt_eval_urls}

\end{table*}

\end{document}